\documentclass[conference]{IEEEtran}

\usepackage{cite}
\usepackage{amsmath,amssymb,amsfonts}
\usepackage{algorithmic}

\usepackage{graphicx}

\usepackage{array}
\usepackage{booktabs}
\usepackage{multirow}

\usepackage{textcomp}
\usepackage{url}

\def\BibTeX{{\rm B\kern-.05em{\sc i\kern-.025em b}\kern-.08em
    T\kern-.1667em\lower.7ex\hbox{E}\kern-.125emX}}

\begin{document}

\title{Parameter-Efficient Fine-Tuning for Low-Resource Languages: A Comparative Study of LLMs for Bengali Hate Speech Detection}

\author{
\IEEEauthorblockN{Akif Islam}
\IEEEauthorblockA{\textit{Department of Computer Science and Engineering} \\
\textit{University of Rajshahi} \\
Rajshahi 6205, Bangladesh \\
iamakifislam@gmail.com}
\and
\IEEEauthorblockN{Mohd Ruhul Ameen}
\IEEEauthorblockA{\textit{College of Engineering and Computer Sciences} \\
\textit{Marshall University} \\
Huntington, WV, USA \\
ameen@marshall.edu}
}

\maketitle

\begin{center}
\vspace{-0.75em}
\textit{This is the authors’ accepted manuscript (preprint) of a paper accepted to IEEE COMPAS 2025. 
The final version will appear in IEEE Xplore.}
\end{center}

\begin{abstract}
Bengali social media platforms have witnessed a sharp increase in hate speech, disproportionately affecting women and adolescents. While datasets such as BD-SHS provide a basis for structured evaluation, most prior approaches rely on either computationally costly full-model fine-tuning or proprietary APIs. This paper presents the first application of Parameter-Efficient Fine-Tuning (PEFT) for Bengali hate speech detection using LoRA and QLoRA. Three instruction-tuned large language models—Gemma-3-4B, Llama-3.2-3B, and Mistral-7B were fine-tuned on the BD-SHS dataset of 50,281 annotated comments. Each model was adapted by training fewer than 1\% of its parameters, enabling experiments on a single consumer-grade GPU. The results show that Llama-3.2-3B achieved the highest F1-score of 92.23\%, followed by Mistral-7B at 88.94\% and Gemma-3-4B at 80.25\%. These findings establish PEFT as a practical and replicable strategy for Bengali and related low-resource languages.
\end{abstract}


\section{Introduction}

By 2025, Bengali is estimated to have over 242 million native speakers and approximately 43 million additional second-language users. Thus, it ranks as the sixth most widely spoken language by native speakers and the seventh most spoken language overall in the world\cite{ethnologue2024}. With the growth of digital platforms, Bengali-speaking communities have seen a sharp increase in online hate speech, which has brought harmful consequences such as cyberbullying and serious mental health challenges \cite{noor2025cyberbullying_mdd_bangladesh}.In January 2025, Bangladesh recorded approximately 60 million social media user identities, representing 34.3\% of the national population\cite{DataReportal2025Bangladesh}. However, alongside this growth, many users have exploited these platforms to engage in harmful behaviors such as public harassment of women, political and religious intimidation, and other forms of online abuse, with women being disproportionately targeted.Studies in Bangladesh reveal the prevalence of online hate speech and its harmful effects on women and adolescents, underscoring the need for effective detection and prevention methods \cite{apc_bangladesh_hate_2022, actionaidbd_ovaw_2022}. Bengali’s unique language features, like complex morphology and heavy code-mixing with English (‘Banglish’), make the task even harder\cite{karim2020classification}. The problem becomes worse with mis-spelled words and sentences carrying double meanings. While high-quality datasets exist for languages like English, Chinese, and German, there continues to be a notable lack of well-curated, large and reliable datasets for Bengali\cite{Lima2023DataModelCentricNER}.

Most earlier work on Bengali hate speech detection has used traditional machine learning, deep learning, or BERT-based fine-tuning \cite{Karim2023_MultimodalHateBengali} , and some rely on large language model APIs\cite{hasan-etal-2024-zero}, which are costly instead of local model training. But when trying to train or fine-tune very large language models (with billions of parameters), consumer-grade hardware often fails because of limited GPU capacity and video random-access Memory (VRAM). Since typical users don’t have access to industry-scale GPUs, we instead adopt a parameter-efficient fine-tuning (PEFT) strategy\cite{Xu2023_PEFTReview}. In this paper, we leverage parameter efficient fine-tuning (PEFT) methods, which can train up to twice as faster while cutting VRAM usage by around 70\%\cite{Han2024_PEFTSurvey,unsloth_framework_2024}. To the best of our knowledge, this is the first study of its kind for Bengali hate speech. We conduct a comparative evaluation on Gemma, Llama, and Mistral backbones, delivering a scalable, resource-efficient, and high-performance solution tailored for Bengali—and put forward this approach as a prototype for other low-resource languages.

\section{Literature Review}

\subsection{Bengali Hate Speech Detection: From Traditional to Transformer Models}
Hate speech detection in Bengali has evolved in line with broader NLP trends. Early work used traditional machine learning with lexical features and classifiers such as SVM and logistic regression \cite{Davidson2017AutomatedHS}, but these underperformed in Bengali due to limited data and morphological complexity. The adoption of deep learning, particularly CNNs and LSTMs with word embeddings, improved accuracy though generalization remained a challenge \cite{maruf2024survey}.

A major advancement was the release of curated Bengali hate speech datasets. Romim et al. introduced BD-SHS\cite{romim2022bdshs}, a benchmark of 50,281 comments enabling systematic evaluation across domains. Earlier, Ishmam and Sharmin collected 5K annotated Facebook comments\cite{ishmam_alvi_hate}, Karim et al. released 8K samples covering multiple categories \cite{karim2021deephateexplainer}, and Das et al. created a 10K dataset incorporating both native-script and Romanized Bengali posts \cite{das2022hate}. These resources highlight the linguistic and sociocultural complexities of Bengali hate speech, particularly code-mixing and orthographic variation.

Transformer-based architectures marked another leap. Fine-tuned BERT and XLM-R consistently outperformed earlier neural models, with F1 scores above 85\% \cite{tanvirularxiv}. Bangla-BERT, trained specifically for Bengali, and multilingual models like MuRIL demonstrated effective transfer from high-resource to low-resource settings \cite{bhattaacl2022hate}. However, the high cost of full fine-tuning limited scalability in resource-constrained environments.

\subsection{Advances with Large and Tiny Language Models}
Sen et al. proposed HateTinyLLM\cite{Sen2024HateTinyLLMH}, demonstrating that compact decoder-only models such as TinyLlama-1.1B, Phi-2, and OPT-1.3B, when fine-tuned, can achieve accuracies above 80\% while being far more resource-efficient than larger language models. Their results emphasize that smaller LLMs, fine-tuned via adapters and LoRA, are viable for low-resource deployment, though their experiments were conducted in English. In contrast, larger models such as GPT-4 and Mixtral 8-7B often demonstrate strong zero and few-shot performance \cite{Bauer2024TRAC}, though their deployment typically entails higher inference cost and engineering complexity due to their substantial model scale \cite{Mixtral2024}.

\subsection{Parameter-Efficient Fine-Tuning Across Languages}
To reduce the cost of adapting billion-parameter models, Parameter-Efficient Fine-Tuning (PEFT) methods have gained prominence. LoRA introduced low-rank adapters, cutting trainable parameters by more than 99\% without accuracy loss \cite{hu2022lora}. QLoRA extended this with 4-bit quantization and optimized backpropagation, enabling adaptation of multi-billion-parameter models on consumer-grade GPUs \cite{dettmers2023qlora}.

Evidence of PEFT’s effectiveness in low-resource languages is growing. Sidibomma et al. applied LoRA to Hindi and Nepali hate speech detection, achieving notable improvements with lower computational cost \cite{sidibomma2025llmsagainsthate}. Their work underscores the viability of PEFT in Devanagari-scripted languages, suggesting direct applicability to Bengali. Adapter-based approaches have also been applied in meta-learning frameworks such as HateMAML\cite{10100717} for low-resource detection tasks \cite{Sen2024HateTinyLLMH}, further confirming the adaptability of PEFT.

\subsection{Research Gap}
Despite extensive progress, no prior study has systematically applied PEFT to Bengali hate speech detection using large-scale benchmarks like BD-SHS. Most Bengali work focuses on traditional or transformer-based fine-tuning \cite{das2022hate, maruf2024survey}. While prior studies in Hindi and Nepali demonstrate the potential of PEFT, its application to Bengali remains unexplored. This gap motivates our work to investigate parameter-efficient and scalable solutions specifically for Bengali.

\section{Methodology}

This work explores parameter-efficient fine-tuning (PEFT) for Bengali hate speech detection by applying Low-Rank Adaptation (LoRA) and its quantized variant (QLoRA) to large language models (LLMs). The experiments were carried out using the Unsloth framework\cite{unsloth_framework_2024}, which enables efficient training of multi-billion-parameter models on a single consumer-grade GPU. The methodology covers dataset preparation, fine-tuning strategy, model configurations, prompting setup, training infrastructure, and evaluation design.

\begin{figure}[!b]
\centering
\includegraphics[width=0.48\textwidth]{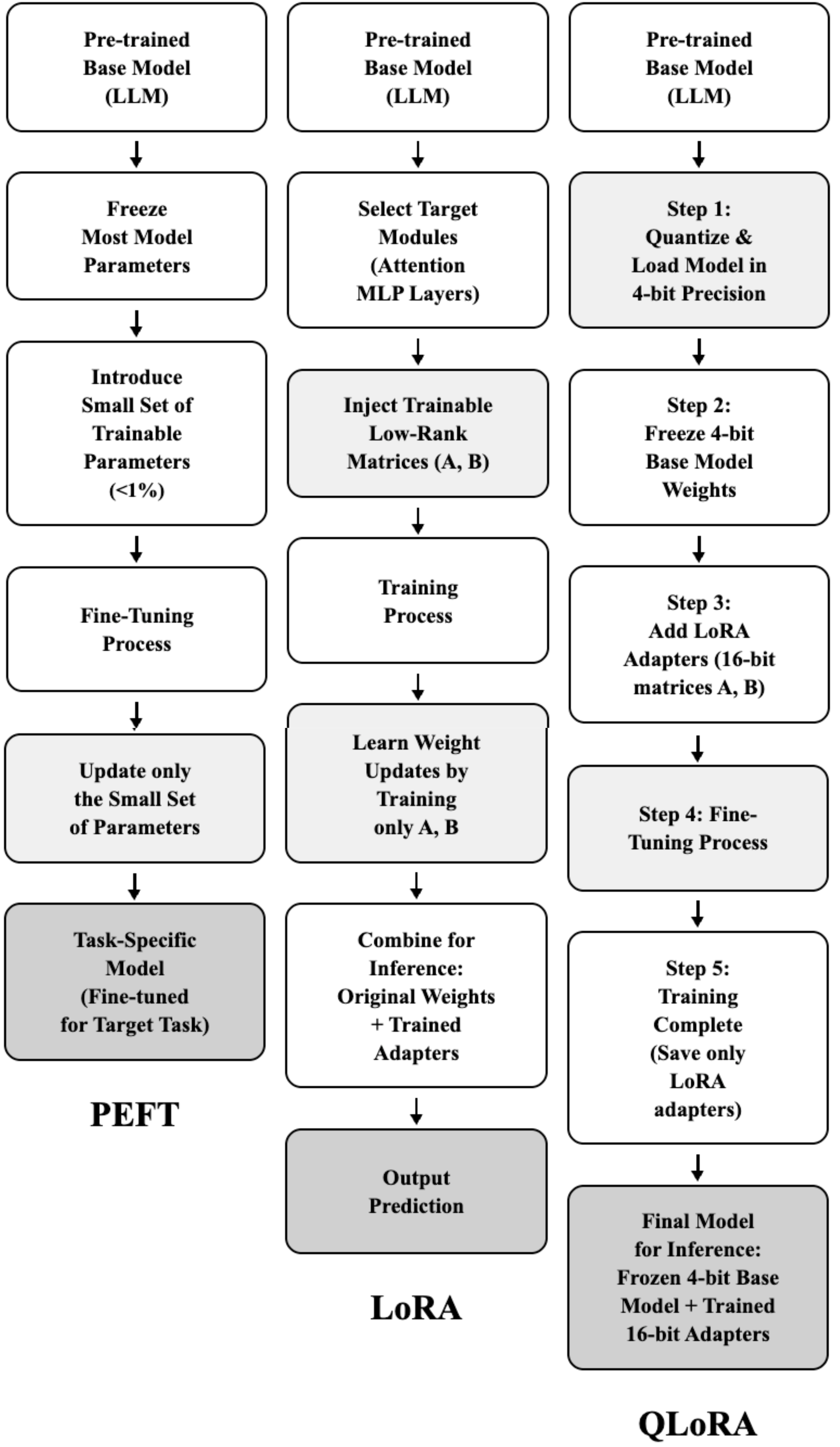}
\caption{Overview of the PEFT approach. LoRA introduces trainable low-rank matrices into transformer layers, while QLoRA further reduces memory usage by quantizing base weights to 4 bits.}
\label{fig:peft_overview}
\end{figure}

\subsection{Dataset}

\begin{table}[!t]
\centering
\caption{Sample Sentences and Hate Speech Labels from BD-SHS Dataset}
\label{tab:hate_speech}
\renewcommand{\arraystretch}{1.1}
\begin{tabular}{c}
\includegraphics[width=0.9\linewidth]{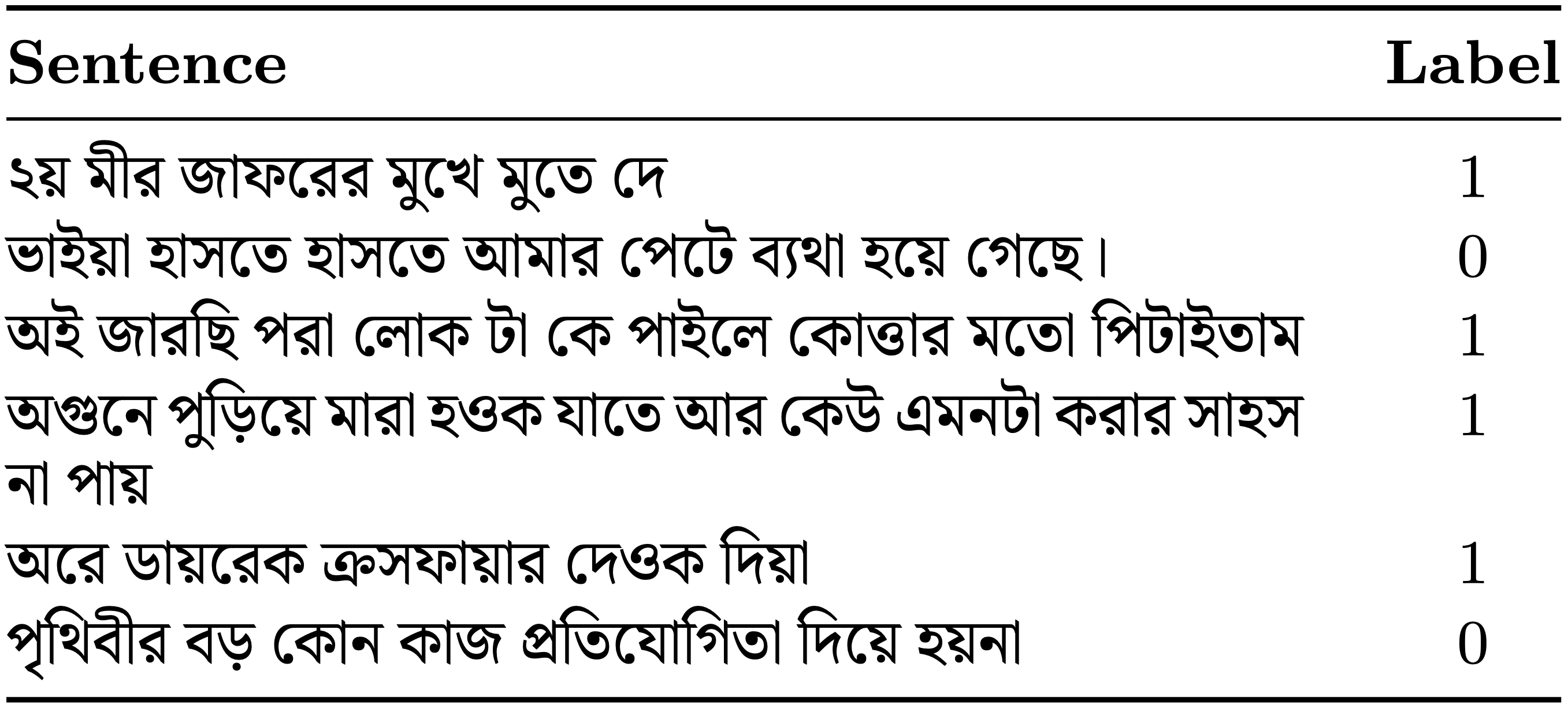}
\end{tabular}
\end{table}

Experiments were conducted on the BD-SHS dataset~\cite{romim2022bdshs}, the largest publicly available benchmark for Bengali hate speech detection. The corpus comprises 50,281 manually annotated comments collected from Facebook, YouTube, and TikTok across domains such as politics, sports, entertainment, religion, crime, and memes. Reliability was ensured through crowdsourced triple labeling followed by expert verification. Minimal preprocessing, limited to removal of non-printing characters and excess whitespace, was applied to preserve the linguistic signal. The dataset is balanced (50\% hate vs.\ 50\% non-hate) and partitioned into training, validation, and test splits in an 80/10/10 ratio. In addition to binary labels, it includes target categories (e.g., individual, male, female, group) and hate types (e.g., slander, religious hate, incitement to violence), supporting robust evaluation of classification models. Representative examples and statistics are summarized in Tables~\ref{tab:hate_speech} and~\ref{tab:dataset_stats}.

\begin{table}[!t]
\centering
\caption{BD-SHS Dataset Statistics}
\label{tab:dataset_stats}
\resizebox{0.95\columnwidth}{!}{%
\begin{tabular}{lcc}
\toprule
\textbf{Attribute} & \textbf{Count/Value} & \textbf{Percentage} \\
\midrule
\multicolumn{3}{c}{\textbf{Dataset Split}} \\
Training Samples & 40,224 & 80.0\% \\
Validation Samples & 5,028 & 10.0\% \\
Test Samples & 5,029 & 10.0\% \\
Total Samples & 50,281 & 100\% \\
\midrule
\multicolumn{3}{c}{\textbf{Class Distribution}} \\
Hate Speech (Label = 1) & 25,141 & 50.0\% \\
Non-Hate Speech (Label = 0) & 25,140 & 50.0\% \\
\midrule
\multicolumn{3}{c}{\textbf{Target Categories}} \\
Individual (ind) & 18,094 & 36.0\% \\
Male & 8,042 & 16.0\% \\
Group & 6,034 & 12.0\% \\
Female & 3,015 & 6.0\% \\
Not Specified & 15,096 & 30.0\% \\
\midrule
\multicolumn{3}{c}{\textbf{Hate Speech Types}} \\
Call to Violence & 12,571 & 25.0\% \\
Slander & 8,042 & 16.0\% \\
Religious Hate & 4,521 & 9.0\% \\
Not Specified & 25,147 & 50.0\% \\
\bottomrule
\end{tabular}
}
\end{table}

\subsection{Parameter-Efficient Fine-Tuning}

\subsubsection{Full Fine-Tuning vs. PEFT}
Conventional full fine-tuning updates every weight and bias in the network, which for modern LLMs can exceed billions of parameters and demand more than 100\,GB of GPU memory\cite{hu2022lora}. Partial fine-tuning strategies that unfreeze selected layers mitigate that cost but remain computationally demanding and prone to overfitting \cite{taehovram}. In contrast, adapter-based methods such as LoRA introduce lightweight, low-rank trainable matrices into transformer layers while keeping the pretrained weights frozen. This reduces the number of trainable parameters to under 1\% of the full model but still enables effective task adaptation. A further key benefit of adapters is modularity: multiple task-specific adapters can be trained and swapped atop the same base model without retraining it from scratch.

\subsubsection{LoRA and QLoRA}
LoRA freezes the pretrained model weights and injects low-rank adapter matrices into each transformer layer where only these adapters are trained. This method significantly reduces memory and compute overhead \cite{hu2022lora}. QLoRA further quantizes the base model to 4-bit precision and backpropagates through it into the frozen model into LoRA adapters, enabling tuning of very large models under constrained memory \cite{dettmers2023qlora}. In our implementation, combining QLoRA with optimized kernels and dynamic memory management reduces GPU memory usage by up to 70\%, and yields nearly double effective throughput compared to unoptimized PEFT pipelines.

\subsection{Model Selection and Configuration}
We fine-tuned three open-source instruction-tuned models: Gemma-3-4B, Llama-3.2-3B, and Mistral-7B. Models exceeding 10B parameters were excluded as infeasible under a 24\,GB GPU budget, while models smaller than 2B parameters were considered insufficient to capture the linguistic complexity of Bengali. While Qwen does include multilingual support (over 29 languages) \cite{Qwen2Report}, its architecture and training emphasis favor Chinese–English scenarios, raising uncertainty about its performance on Bengali text. Similarly, Phi models \cite{Phi4TechnicalReport} are more English-centric, with less published evidence of robust adaptation to low-resource languages. In contrast, the Gemma family \cite{gemma2024} is explicitly documented as multilingual across supported models, and Llama and Mistral are widely adopted in community workflows with demonstrated cross-lingual transfer. 

For all models, we applied LoRA with rank = 16, $\alpha = 16$, and learning rate $2\times10^{-4}$, initializing them in 4-bit quantization mode to enable QLoRA. Only the adapter matrices were fine-tuned. In every case, fewer than 1\% of total parameters were updated, underscoring the parameter efficiency of PEFT under constrained hardware.

\subsection{Instruction Prompting}
To align the models with the binary classification objective, we integrated instruction-based prompting directly within the fine-tuning process pipeline. A single Bengali prompt template was used consistently during training, ensuring that models always receive the same structured input format. Example templates are shown in Table~\ref{tab:prompt_templates}.

\begin{table}[!t]
\centering
\caption{Instruction Prompt Templates for Bengali Hate Speech Detection}
\label{tab:prompt_templates}
\renewcommand{\arraystretch}{1.15}
\begin{tabular}{c}
\includegraphics[width=0.9\linewidth]{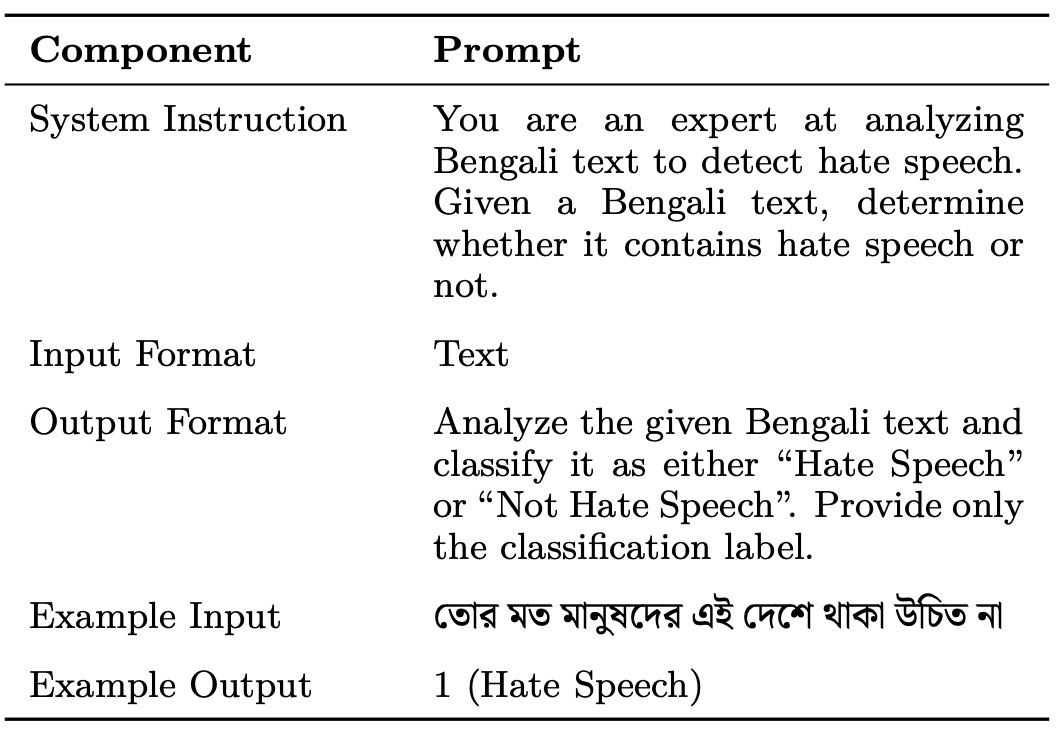}
\end{tabular}
\end{table}

\subsection{Experimental Setup and Evaluation}

All experiments were executed on a single NVIDIA RTX 4090 GPU with maxium sequence length of 2048. Batch sizes scaled with model: size 32 for Gemma-3-4B, 8 for Llama-3.2-3B, and 4 for Mistral-7B, with gradient accumulation to offset memory constraints. Model performance was primarily evaluated via the weighted F1-score, capturing both false positives and false negatives. Accuracy was also reported but considered less informative in hate speech detection due to its equal treatment of error types. The F1-score is defined as
\[
\mathrm{F1\mbox{-}Score} = 2 \cdot \frac{\mathrm{Precision} \cdot \mathrm{Recall}}{\mathrm{Precision} + \mathrm{Recall}}.
\]
Confusion matrices supported error analysis, while runtime and GPU memory usage were tracked for efficiency. Preprocessing, hyperparameters, and dataset splits were kept consistent to ensure fair and reproducible comparison.

\section{Results and Discussion}

\subsection{Parameter-Efficient Fine-Tuning Performance}
This section reports a structured evaluation of Parameter-Efficient Fine-Tuning (PEFT) for Bengali hate speech detection using large language models. Table~\ref{tab:llm_results} summarizes performance across the three of our fine-tuned models, while Table~\ref{tab:model_comparison} shows predicted outputs on sample inputs. On the BD-SHS benchmark, Llama-3.2-3B achieved the highest F1-Score of 92.23\%, followed by Mistral-7B with 88.94\% and Gemma-3-4B with 80.25\%.

\begin{table}[!b]
\centering
\caption{PEFT-based LLM performance on BD-SHS dataset}
\label{tab:llm_results}
\small
\renewcommand{\arraystretch}{1.15}
\resizebox{\columnwidth}{!}{%
\begin{tabular}{lccccc}
\toprule
\textbf{Model} & \textbf{Acc} & \textbf{Prec} & \textbf{Rec} & \textbf{F1} & \textbf{Mem} \\
\midrule
Gemma-3-4B     & 81.31\% & 81.28\% & 79.25\% & 80.25\% & 15.6 GB \\
\textbf{Llama-3.2-3B} & \textbf{92.48\%} & \textbf{93.00\%} & \textbf{92.00\%} & \textbf{92.23\%} & \textbf{12.8 GB} \\
Mistral-7B     & 88.94\% & 89.12\% & 88.76\% & 88.94\% & 14.2 GB \\
\bottomrule
\end{tabular}
}
\end{table}

\begin{figure}[!b]
\centering
\includegraphics[width=0.4\textwidth]{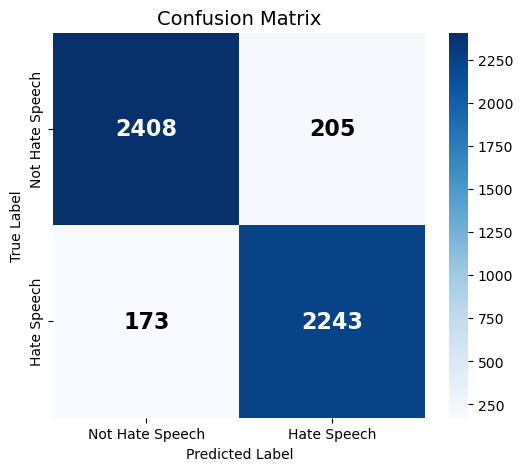}
\caption{Confusion matrix of Llama-3.2-3B on the BD-SHS test set}
\label{fig:llama_confusion}
\end{figure}

\begin{table}[!t]
\centering
\caption{Predicted Output of Each Model alongside Actual Ground Truth for Sample Inputs}
\label{tab:model_comparison}
\renewcommand{\arraystretch}{1.15}
\begin{tabular}{c}
\includegraphics[width=0.9\linewidth]{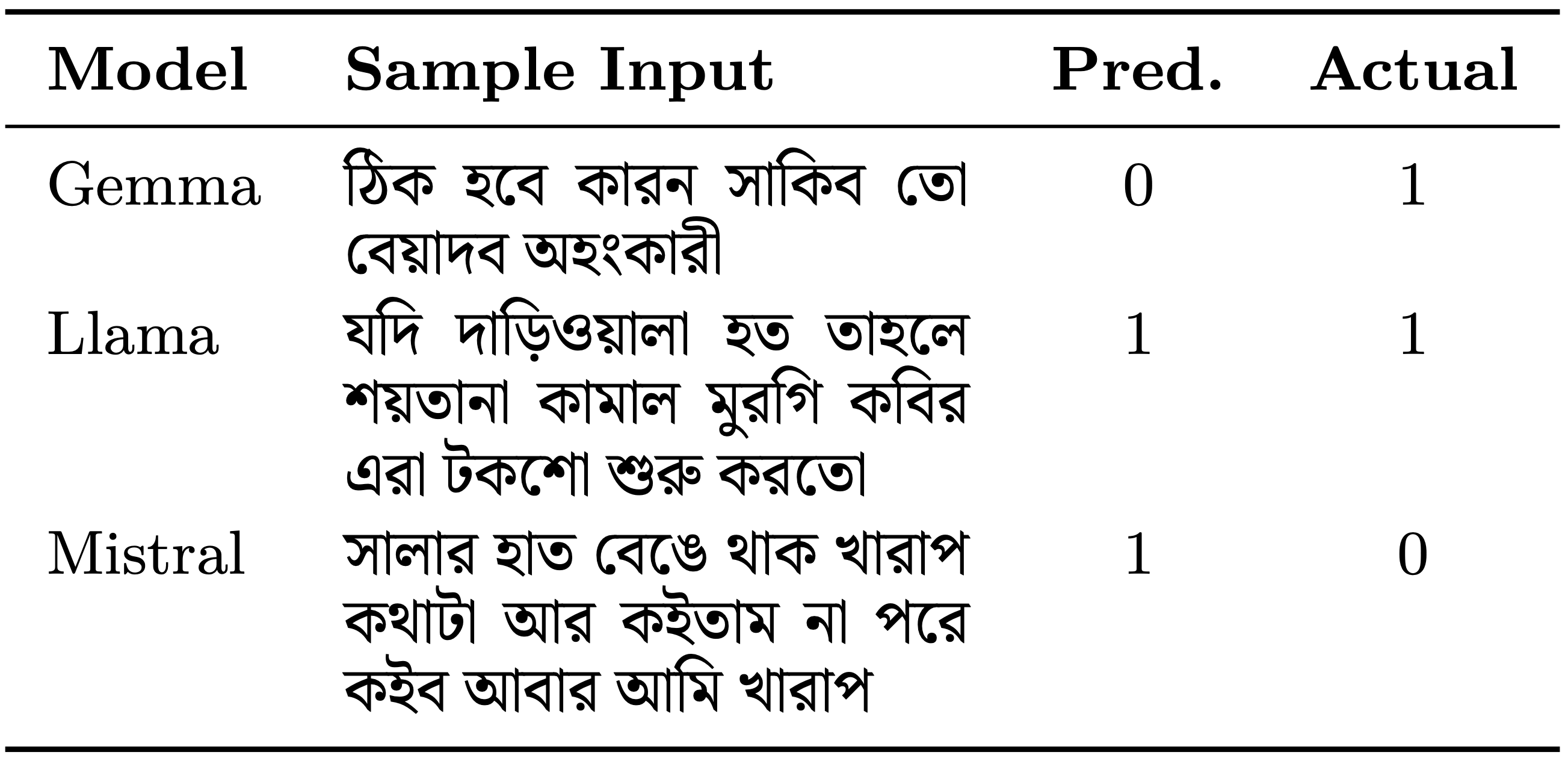}
\end{tabular}
\end{table}

\subsection{Individual Model Analysis and Comparative Insights}

When fine-tuned on BD-SHS, \textbf{Llama-3.2-3B} achieved an F1-Score of 92.23 \% and accuracy of 92.48 \%. Its confusion matrix (Fig.~\ref{fig:llama_confusion}) and precision–recall balance (Table~\ref{tab:llm_results}) suggest reliable performance on both classes. This result may partly reflect architectural features such as RoPE positional embeddings, SwiGLU activations, RMSNorm normalization, and a multilingual subword tokenizer that supports generalization \cite{zhao2023surveyllm}.

The \textbf{Mistral-7B} model obtained an F1-Score of 88.94 \%. Mistral uses grouped-query attention (GQA) and sliding window attention (SWA) to improve inference efficiency \cite{jiang2023mistral7b}, though these benefits may be reduced in short social media settings. Its larger parameter count also increases memory demand, which can exacerbate overfitting in low-data regimes.

\textbf{Gemma-3-4B} obtained an F1-Score of 80.25 \%. It provides a lighter computational footprint and faster convergence, though its tokenizer and representational power may be less optimal for Bengali relative to larger models. Architectural surveys note that Gemma shares many design elements with LLaMA but diverges in activation functions and tokenization strategies \cite{zhao2023surveyllm}.

These findings indicate that simply increasing parameter count or adding advanced attention mechanisms is not enough for low-resource language tasks. What matters more is how well the model’s architecture, tokenizer, and parameter-efficient fine-tuning work together to deliver strong performance on the target task.

\begin{table}[h]
\centering
\caption{Performance comparison on hate speech detection across studies.}
\label{tab:comparison_bdshs}
\resizebox{\columnwidth}{!}{
\begin{tabular}{l l c c}
\toprule
\textbf{Study} & \textbf{Approach} & \textbf{Best Model} & \textbf{F1-Score} \\
\midrule
Karim et al. (2021) \cite{karim2021deephateexplainer} & Conv-LSTM (DL) & Conv-LSTM & 78.0\% \\
\textbf{This Work} & \textbf{PEFT-LLM} & \textbf{Gemma-3-4B} & \textbf{80.25\%} \\
Karim et al. (2021) \cite{karim2021deephateexplainer} & Transformer (BERT) & XLM-RoBERTa & 82.0\% \\
Sen et al. (2024) \cite{Sen2024HateTinyLLMH} & PEFT (LoRA) & OPT-1.3B & 83.0\% \\
Sidibomma et al. (2025) \cite{das2022hate} & PEFT (LoRA) & Phi-3-medium &  88.91\% \\
\textbf{This Work} & \textbf{PEFT-LLM} & \textbf{Mistral-7B} & \textbf{88.94\%} \\
Sidibomma et al. (2025) \cite{sidibomma2025llmsagainsthate} & PEFT (LoRA) & Nemo & 90.05\% \\
Romim et al. (2022) \cite{romim2022bdshs} & BiLSTM (DL) & BiLSTM+IFT & 91.0\% \\
\textbf{This Work} & \textbf{PEFT-LLM} & \textbf{Llama-3.2-3B} & \textbf{92.23\%} \\
\bottomrule
\end{tabular}
}
\end{table}

% \begin{table}[h]
% \centering
% \caption{Performance Comparison on Hate Speech Detection}
% \label{tab:comparison_bdshs}
% \resizebox{\columnwidth}{!}{
% \begin{tabular}{|l|l|c|c|}
% \hline
% \textbf{Study} & \textbf{Approach} & \textbf{Best Model} & \textbf{F1-Score} \\
% \hline
% Karim et al. (2021) \cite{karim2021deephateexplainer} & Conv-LSTM (DL) & Conv-LSTM & 78.0\% \\
% \hline
% \textbf{This Work} & \textbf{PEFT-LLM} & \textbf{Gemma-3-4B} & \textbf{80.25\%} \\
% \hline
% Karim et al. (2021) \cite{karim2021deephateexplainer} & Transformer (BERT) & XLM-RoBERTa & 82.0\% \\
% \hline
% Sen et al. (2024) \cite{Sen2024HateTinyLLMH} & PEFT (LoRA) & OPT-1.3B & 83.0\% \\
% \hline
% Das et al. (2022) \cite{das2022hate} & Transformer (BERT) & Bangla-BERT & $\sim$85.0\% \\
% \hline
% \textbf{This Work} & \textbf{PEFT-LLM} & \textbf{Mistral-7B} & \textbf{88.94\%} \\
% \hline
% Sidibomma et al. (2025) \cite{sidibomma2025llmsagainsthate} & PEFT (LoRA) & Nemo (6.97B) & 90.05\% \\
% \hline
% Romim et al. (2022) \cite{romim2022bdshs} & ML Baseline & Custom Embedding & 91.0\% \\
% \hline
% \textbf{This Work} & \textbf{PEFT-LLM} & \textbf{Llama-3.2-3B} & \textbf{92.23\%} \\
% \hline
% \end{tabular}
% }
% \end{table}

\begin{figure}[!t]
\centering
\includegraphics[width=0.5\textwidth]{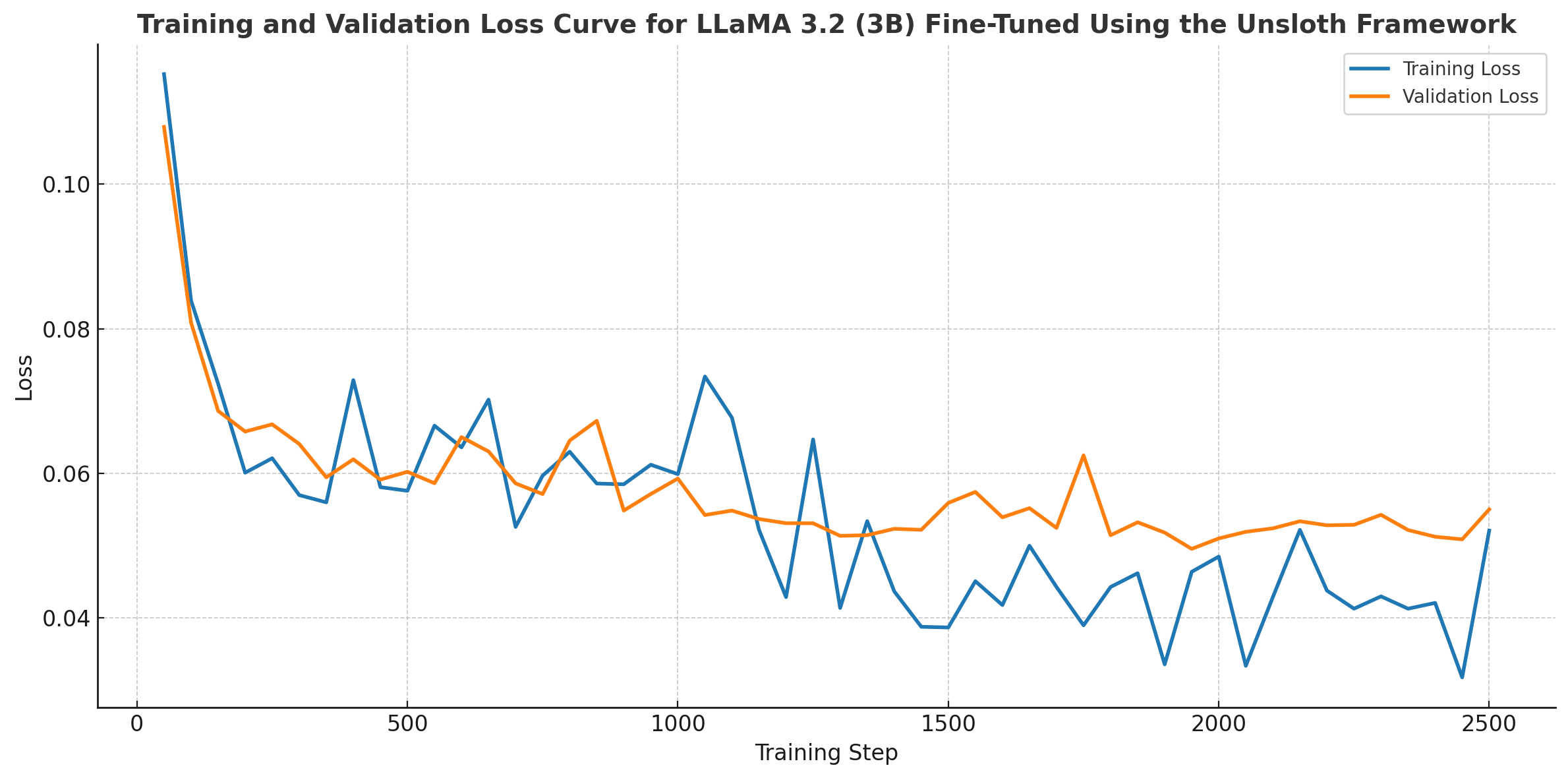}
\caption{Training and Validation Loss Curves for Llama-3.2-3B Fine-tuning. The curves demonstrate stable convergence without overfitting over 3 epochs, with final training loss of 0.145 and validation loss of 0.162.}
\label{fig:llama_loss}
\end{figure}

\subsection{Novel Contributions and Significance}
This work provides the first systematic evidence that PEFT with instruction-tuned open LLMs can deliver state-of-the-art performance for Bengali hate speech detection while updating fewer than 1\% of model parameters. The approach reduces the practical barrier to high-quality moderation models in low-resource settings and offers a replicable template for related Indic and underrepresented languages.

\subsection{Computational Efficiency Analysis}
Table~\ref{tab:efficiency_analysis_corrected} summarizes computational efficiency. Across models, the trainable fraction remains below 1\% (Gemma-3-4B: 0.69\%, Llama-3.2-3B: 0.75\%, Mistral-7B: 0.58\%), and memory reductions of 65–75\% are observed under the quantized PEFT setup. These figures indicate that the performance gains reported above are attainable under practical memory budgets, supporting the case for PEFT in deployment-oriented scenarios.

\begin{table}[!t]
\centering
\caption{Computational Efficiency Comparison}
\label{tab:efficiency_analysis_corrected}
\small
\renewcommand{\arraystretch}{1.15}
\resizebox{\columnwidth}{!}{%
\begin{tabular}{lcccc}
\toprule
\textbf{Model} & \textbf{Total Params} & \textbf{Trainable Params} & \textbf{Efficiency} & \textbf{Mem. Red.} \\
\midrule
Gemma-3-4B    & 4.24B & 29.8M (0.69\%) & 142$\times$ & 70\% \\
Llama-3.2-3B  & 3.20B & 24.3M (0.75\%) & 132$\times$ & 65\% \\
Mistral-7B    & 7.00B & 41.9M (0.58\%) & 167$\times$ & 75\% \\
\bottomrule
\end{tabular}%
}
\end{table}

\subsection{Key Findings and Practical Deployment Considerations}

Performance on BD-SHS depends not only on model size but also on how well architecture and tokenizer design align with the task. Llama-3.2-3B achieves the best balance of precision and recall, generalizing well to code-mixed and noisy text. Mistral-7B is competitive but less adapted to BD-SHS’s variation, while Gemma-3-4B offers a lightweight option with a consistent F1-score trade-off. For deployment, Llama-3.2-3B is recommended for maximizing F1-score on BD-SHS-like content. Mistral-7B may be preferred for longer or more demanding inputs, whereas Gemma-3-4B is suitable when memory efficiency is the priority. Memory footprints for each model are summarized in Table~\ref{tab:llm_results}.

\section{Conclusion}
This paper introduced the first study to apply Parameter-Efficient Fine-Tuning (PEFT) for Bengali binary hate speech detection. Future research should address more nuanced challenges such as dialectal variation, non-social media domains, and robustness under explicit code-mixing conditions to better reflect real-world moderation scenarios. Using LoRA and QLoRA on large instruction-tuned models, we showed that effective moderation systems can be built even under limited computational resources. The results suggest that careful alignment between architecture, tokenizer behavior, and fine-tuning strategy matters more than simply increasing model size. In future, access to larger datasets, broader domain coverage, and stronger hardware will allow even better performance. With the rapid arrival of new LLMs, combining them with PEFT offers a practical path to strengthen hate speech detection in Bengali and to extend similar advances to other low-resource languages.

\bibliographystyle{IEEEtran} % Use IEEE style
\bibliography{references}    
% Refers to references.bib file (no extension)
% \input{main.bbl}

\end{document}